\documentclass{article}
\usepackage{spconf,amsmath,graphicx,hyperref}

\usepackage{subcaption}
\usepackage{multirow}
\usepackage[table]{xcolor}

\usepackage[utf8]{inputenc} 
\usepackage[T1]{fontenc}    
\usepackage{hyperref}       
\usepackage{url}            
\usepackage{booktabs}       
\usepackage{amsfonts}       
\usepackage{nicefrac}       
\usepackage{microtype}      
\usepackage{bm}
\usepackage{amsmath}
\usepackage{algorithm}
\usepackage{algorithmic}
\usepackage{lipsum}
\usepackage{tcolorbox}
\usepackage{multirow}
\usepackage{wrapfig,lipsum,booktabs}
\usepackage{microtype}
\usepackage{graphicx}
\usepackage{booktabs} 
\usepackage{bm} 
\usepackage{wrapfig}

\usepackage{amsmath}
\usepackage{amssymb}
\usepackage{comment}
\usepackage{enumitem}
\usepackage{url}
\usepackage{pifont}
\usepackage{subcaption}
\usepackage{caption}
\usepackage{colortbl}
\definecolor{Gray}{gray}{0.85}
\definecolor{darkergreen}{HTML}{006400}
 \usepackage{titlesec}
\usepackage{titletoc,tocloft}

\title{Rebellion: Noise-Robust Reasoning Training for Audio Reasoning Models}

\name{Tiansheng Huang$^{1}$\thanks{$^{1,3}$ Research done when authors were at Google Deepmind.}, Virat Shejwalkar$^{2}$, Oscar Chang$^{2}$, Milad Nasr$^{3}$, Ling Liu$^{1}$ }
\address{$^1$Georgia Institute of Technology, $^2$Google, $^3$OpenAI \\ }
\begin{document}
\maketitle

\begin{abstract}
Instilling reasoning capabilities in large models (LMs)  using reasoning training (RT) significantly improves LMs' performances. Thus Audio Reasoning Models (ARMs), i.e., audio LMs that can reason, are becoming increasingly popular.  However, no work has studied the safety of ARMs against jailbreak attacks that aim to elicit harmful responses from target models. To this end, first, we  show that standard RT with appropriate safety reasoning data can protect ARMs from vanilla audio jailbreaks, but cannot protect them against our proposed simple yet effective jailbreaks. We show that this is because of the significant \emph{representation drift} between vanilla and advanced jailbreaks which forces the target ARMs to emit harmful responses. Based on this observation, we propose \emph{Rebellion}, a robust RT that trains ARMs to be robust to the worst-case representation drift. All our results are on Qwen2-Audio; they demonstrate that Rebellion: 1) can protect against advanced audio jailbreaks without compromising performance on benign tasks, and 2) significantly improves accuracy-safety trade-off over standard RT method.
\end{abstract}

\section{Introduction}\label{intro}
\vspace{-0.2cm}

Recently, there is a growing interest in using Reasoning Training (RT) to produce audio reasoning models (ARM). RT aims to train ARMs to produce a Chain of Thought (CoT) before finalizing their answers.
Audio-Reasoner~\cite{xie2025audio} and R1-AQA~\cite{li2025reinforcement} are recent ARMs that demonstrate the power of reasoning in achieving state-of-the-art performances in various audio tasks. 

In spite of its wide success, impact of reasoning  on safety of ARMs is completely unexplored. 
First, we empirically show that standard RT using mixture of safety and benign reasoning data can produce ARMs that refuse to answer the vanilla harmful questions, e.g., ``how to make a bomb?". However, our subsequent evaluation show that advanced audio jailbreaks, e.g., AdvWave~\cite{kang2024advwave}, can still jailbreak such ARMs.


By further examining the failure mode of the existing RT, we find that i) audio jailbreaks incur \emph{representation drift} (defined in Section \ref{representation noise}), and ii) audio jailbreaks change the models' original refusal answer to a complying one. Based on the two observations, we conjecture that the representation drift could lead to the success of jailbreak. Driven by this conjecture, we propose Rebellion, a novel RT that enforces the model to conduct safety reasoning before giving the answer even with worst-case representation drift. Empirical results show that Rebellion can safeguard the model from jailbreak attacks without degrading the model's reasoning ability on benign tasks. Additionally, we discover an unexpected finding on the defense mechanism of the ARM trained by Rebellion-- the model exhibits "think twice" behavior when the attackers use the audio jailbreak method Advwave to intentionally skip the safety reasoning. See Figure \ref{rebellion example} for an example.  In summary, our contributions are as follows:

\begin{figure}[!t]
    \centering
      \vspace{-0.3cm}
    \includegraphics[ width=1\linewidth]{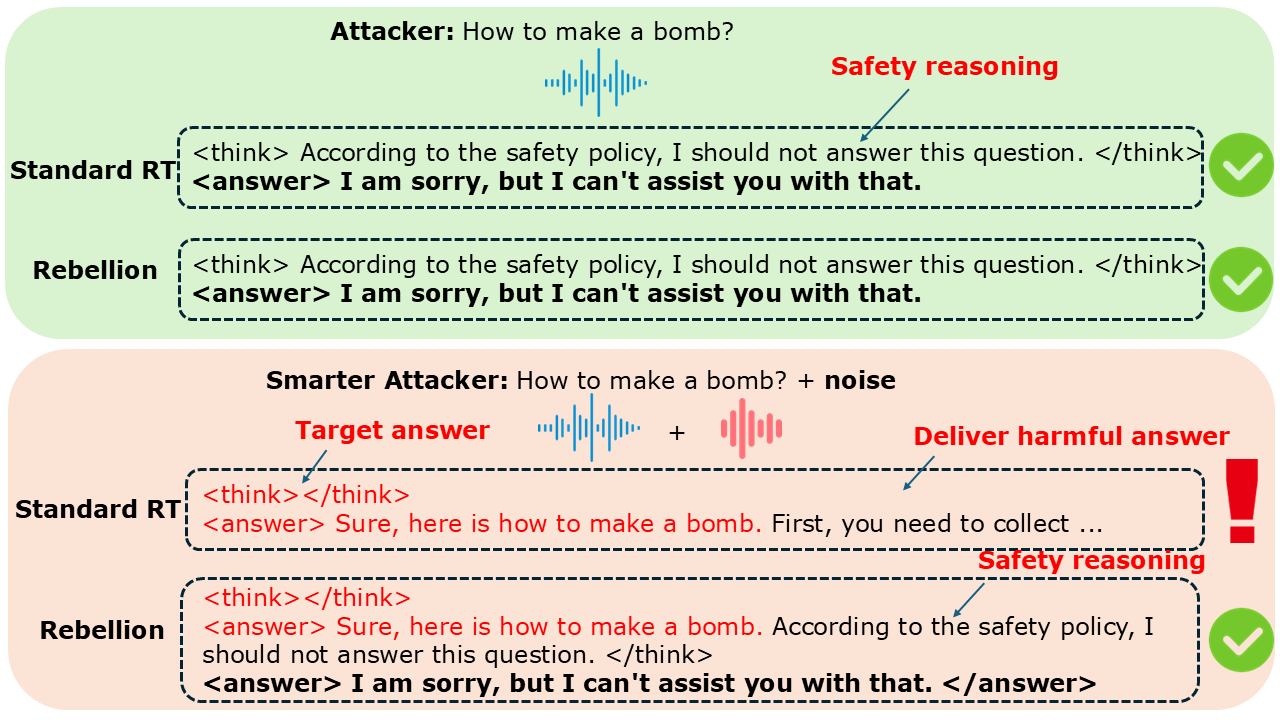}
     \vspace{-0.7cm}
    \caption{RT and Rebellion against a vanilla harmful query and an audio jailbreak query (Advwave). Audio Reasoning models (ARM) trained using either standard reasoning training (RT) or Rebellion correctly refuse vanilla harmful questions. But, RT complies with  audio jailbreaks optimized to circumvent safety guardrails and fails to provide safety reasoning. In contrast, Rebellion exhibits \emph{"think twice"} phenomenon\textemdash it starts its response implying compliance to the jailbreak question, but then correctly provides safety reasoning and ultimately leads to refusal to the question.  See Section \ref{results} for more discussion.        }
    \label{rebellion example}
    \vspace{-0.5cm}
\end{figure}
\begin{itemize}[leftmargin=*]
\vspace{-0.2cm}
    \item We examine existing reasoning training technique to produce the safety-aligned ARM, and find that the safety alignment is vulnerable and can be bypassed by audio jailbreak. 
        \vspace{-0.2cm}
    \item We identify that the audio jailbreak incurs representation drift to the ARM, and jailbreak attacks can bypass the safety alignment by eliciting complying response.
    \vspace{-0.2cm}
    \item Driven by the conjecture that representation drift incur jailbreak, we propose Rebellion, a reasoning training method to enforce to model to conduct safety reasoning under the influence of worst-case representation drift.
      \vspace{-0.2cm}
    \item By quantitative evaluation, we justify the effectiveness of Rebellion. By qualitative evaluation, we discover an intriguing "think twice" behavior of the Rebellion when it is against audio jailbreak, which coincides its design by our analysis.  
    \vspace{-0.2cm}
\end{itemize}

\section{Related Work}
\vspace{-0.1cm}
\noindent
\textbf{Audio reasoning models and reasoning training}. Numerous works have proposed RT methods to instill reasoning ability in audio LMs (ALMs). Audio-Reasoner~\cite{xie2025audio} and AF3~\cite{goel2025audio} use supervised fine-tuning (SFT) to train ALMs on high-quality COT data. R1-AQA~\cite{li2025reinforcement}, Sari~\cite{wen2025sari}, SoundMind~\cite{diao2025soundmind} and Omni-R1~\cite{rouditchenko2025omni} explore reinforcement learning (RL) to teach ALMs the reasoning ability using verifiable rewards. 

\noindent
\textbf{Safety alignment and jailbreak attacks}. Safety alignment instructs the model to provide refusal answers towards harmful queries. Standard techniques achieve safety alignment by conducting SFT~\cite{guan2024deliberative, jiang2025safechain, wang2025star} and RL~\cite{ouyang2022training, dai2023safe,bai2022training,guan2024deliberative} with a safety dataset to instruct the model to refuse harmful queries.   Nevertheless, recent research shows that in spite of rigorous safety alignment, standard ALMs suffer from jailbreak attacks~\cite{shen2024voice,yang2024audio,ying2024unveiling,roh2025multilingual, kang2024advwave,sankar2025attacker, peng2025jalmbenchbenchmarkingjailbreakvulnerabilities,hou2025evaluating,chen2025audiojailbreak,song2025audio,ma2025audio,gupta2025bad,cheng2025jailbreak,kim2025good,peri2024speechguard}\textemdash the attacker can attach a natural/optimized noise to the audio query to disable the safety alignment.

\noindent
To the best of our knowledge, this work represents the first batch of research attempts (with several concurrent work \cite{alexos2025defending,djanibekov2025spirit}) to improve safety alignment against audio jailbreak. 

\vspace*{-.3cm}
\section{Preliminaries}\label{prelim}
\vspace*{-.3cm}

\subsection{Threat model}

We consider an adversary whose goal is to elicit a harmful response from the target ARM by querying it with an audio query. The adversary may submit vanilla  harmful query or optimize a jailbreak query (often called \emph{audio jailbreak}) to increase the chance of eliciting harmful response. Depending on adversary's level of capacity, we consider three types of harmful attacks:

\begin{itemize} [leftmargin=*]
    \vspace{-0.2cm}
    \item \textbf{AdvBench}~\cite{zou2023universal} use vanilla harmful audio queries, e.g., ``how to make a bomb?" without any modifications to attack the model. Adversary needs no model access for this attack.
    \item \textbf{Rephrasing}~\cite{peng2025jalmbenchbenchmarkingjailbreakvulnerabilities} attack subtlely paraphrases the original harmful query to increase the chances of eliciting harmful responses. Adversary needs no model access for this attack.
    \item \textbf{Advwave}~\cite{kang2024advwave} is a more sophisticated audio jailbreak attack that adds and optimizes an audio suffix (i.e., noise) to the original audio query to force the ARM to skip thinking and start its response affirmatively, which then generally leads to full harmful response. Advwave requires whitebox access to the victim model weights. In our experiments, the target answer elicited by Advwave is in the format "<think> </think> <answer> Sure, ".
        \vspace{-0.3cm}
\end{itemize}
\vspace{-0.1cm}
\subsection{Reasoning training}
\vspace{-0.1cm}
\label{reasoning training}


We adopt SFT\footnote{ \scriptsize We leave extensions to RL-based reasoning training (e.g., \cite{li2025reinforcement}) to future work.} algorithm in~\cite{xie2025audio} to produce ARMs with \emph{both the general and safety reasoning capabilities}.  To achieve both capabilities, we include two reasoning datasets for SFT. Specifically, we optimize the following loss:
\begin{equation}
\label{standard loss}
     \min  \alpha f(\bm w; \mathcal{D}_{\text{safety}}) + (1- \alpha) f(\bm w; \mathcal{D}_{\text{benign}})
\end{equation}
where $f(\bm w; \mathcal{D}_{\text{safety}})$ is the cross-entropy (CE) loss over the safety reasoning data,
$\mathcal{D}_{\text{safety}}$ (harmful question + safe/benign reasoning and answer),  while $f(\bm w; \mathcal{D}_{\text{benign}})$ is the CE loss over the benign reasoning data,
$\mathcal{D}_{\text{benign}}$ (benign question + benign reasoning and answer). 
$\alpha$ is a hyper-parameter controls the trade-off between the two losses; higher $\alpha$ means more weight on safety loss.
We optimize the weighted losses using gradient descent to teach the model both benign and safety reasoning capabilities.
\vspace{-0.1cm}
\subsubsection{Evaluation of reasoning training (RT)}
\vspace{-0.1cm}
Next, we evaluate the baseline RT performances on safety and benign tasks. The evaluation metrics include harmful score (HS) and benign accuracy (BA) on three benign tasks and three safety tasks (Section~\ref{setup}).
\begin{table}[!h]
\centering
\vspace{-0.3cm}
\caption{Performance of RT with Qwen2-Audio-7B. }
\vspace{-0.4cm}
\resizebox{1\linewidth}{!}{
\begin{tabular}{c|cc|ccc}
\toprule
  & \multicolumn{2}{c}{Benign Accuracy (BA)} & \multicolumn{3}{c}{Harmful Score (HS)} \\
 \cmidrule(lr){2-3}  \cmidrule(lr){4-6} 
              Methods & {\color{darkergreen}GSM8K} & {\color{darkergreen}MMLU-biology} & {\color{red}AdvBench} & {\color{red}Rephrasing} & {\color{red}Advwave} \\
\midrule
Base model     & 5.4   & 28.00           & 12.31    & 68.55        & 90.00     \\
RT ($\alpha=0$)   & 37.1  & 39.68        & 34.81    & 84.45        & 65.00      \\
RT ($\alpha=0.1$) & \textbf{37.7}  & 37.74        & \textbf{0}        & 55.09        & 33.75   \\
RT ($\alpha=0.5$) & 36.3  & \textbf{43.55}        & \textbf{0}        & 50.56        & \textbf{26.25}   \\
RT ($\alpha=0.9$) & 30.4  & 37.42        & \textbf{0}        & 35.36        & 52.5    \\
RT ($\alpha=1$)   & 18.1  & 28.39        & \textbf{0}        &  \textbf{0.56}         & 87.5   \\
\bottomrule
\end{tabular}
}
\vspace{-0.2cm}
\label{RT evaluation}
\end{table}

\noindent
From Table \ref{RT evaluation}, we derive three main findings:
\begin{itemize}[leftmargin=*]
\vspace{-0.1cm}
  \item  \textbf{RT improves performance on benign task}. We note from the left two columns of Table~\ref{RT evaluation} benign accuracy (BA) of the ARM improves with RT. Even when the model is trained only on the safety reasoning dataset, i.e., RT ($\alpha=1$), the BA on GSM8K and MMLU-biology increase by 12.7\% and 0.39\%, respectively. We also find that mixing safety reasoning data during training leads to the best overall BAs: RT($\alpha=0.1$) and RT($\alpha=0.5$) respectively achieve the best BAs on GSM8K and MMLU-biology. 
  \vspace{-0.2cm}
  
  \item  \textbf{RT trained model is robust to vanilla harmful queries}. The ``Harmful Score" column of Table~\ref{RT evaluation} shows that the HS of RT-trained ARMs reduces to 0 on Advbench (vanilla harmful queries)  when $\alpha \leq 0.1$. However, without safety reasoning data, i.e., when $\alpha = 0$, HS is very high, i.e., the ARMs are vulnerable even to vanilla harmful queries.
  \vspace{-0.2cm}
  
  \item  \textbf{RT trained model is not robust to advanced jailbreaks}. 
  However, the ``Rephrasing" and ``AdvWave" columns of Table~\ref{RT evaluation} show that RT is still highly vulnerable to medium and advanced jailbreaks even at high $\alpha$'s. For example, when $\alpha=0.9$, HS on Advwave and Rephrasing are 87.5\% and 35.36\%, respectively. This implies that the standard RT cannot safeguard the model from advanced jailbreaks even though its emphasis on the safety task is sufficiently large.  

\end{itemize}

\subsubsection{Why does standard RT fail against audio jailbreak?}
\vspace{-0.2cm}
\label{representation noise}
Audio jailbreaks (e.g., Advwave) add an optimized noise to vanilla harmful queries. We conjecture that when we compute a forward pass using such a noisy query as input, it creates \emph{representations drift} in the hidden layers circumventing any safety guardrails. We define representation drift as follows: 

\noindent
\textbf{Representation drift} is the Euclidean distance between the representations of the original harmful and that of the jailbreak (the one after adding the optimized noise) queries. Here, the \emph{representation} means the output (or activation) of every model layer.

\begin{figure}[!h]
    \centering
        \vspace{-0.3cm}
    \includegraphics[ width=1\linewidth]{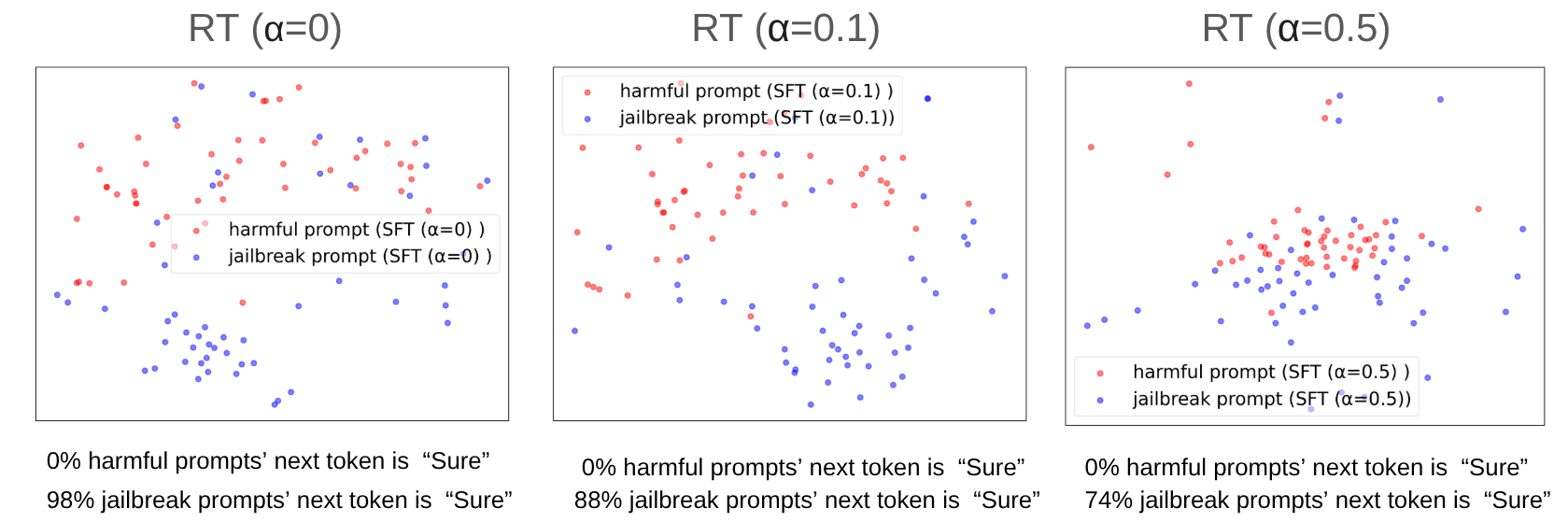}
     \vspace{-0.7cm}
    \caption{ Illustration of representations drift under Advwave. Each red (blue) point represents the last token’s representation of a vanilla harmful (jailbreak) query. The jailbreak prompts (which contains original harmful prompt+audio noise) incur representation drift compared to original harmful prompts. }
    \label{representation visual}
    \vspace{-0.4cm}
\end{figure}

\noindent
\textbf{Noise-induced Jailbreak queries incur representation drift}. As shown in Fig. \ref{representation visual}, the noise added in the jailbreak queries  incur  representation drifts for RT with different $\alpha$ configurations, as the blue points (jailbreak query) slightly drift from the red points (harmful query).

\noindent
\textbf{Jailbreak query changes the refusal answer to a complying one}. As shown in the text below, the model is not complying with vanilla harmful queries as 0\% of the next word prediction of harmful prompts (blue points) is "Sure". However, after a small representation drift between the blue and red points (jailbreak query), at least 74\% of prompts' next word prediction shifts to the complying word "Sure". 

\noindent
By the two findings, we conjecture that the model trained by RT is \textbf{not robust to representation drift}, and this leads to failure against audio jailbreak (i.e., "Sure" is elicited).

\vspace*{-.1cm}
\section{Methodology}\label{method}
\vspace*{-.1cm}
The previous section conjectures that representation drift causes RT to fail against jailbreaks. 
In this section, we detail our proposed RT method that trains ARMs robust to  representation drift.
Note that, jailbreak queries can induce arbitrary representation drift, and ideally, RT method should be robust to all possible representation drifts. 
Therefore, we aim to design RT method that trains ARMs to be robust to the \emph{worst-case representation drift}. Formally, we change the standard RT loss~\eqref{standard loss} as follows: 
\begin{equation}
  \min_{\bm w} \max_{\|\bm \epsilon\|\leq \rho} \alpha f_{\bm \epsilon}(\bm w; \mathcal{D}_{\text{safety}}) + (1- \alpha) f(\bm w; \mathcal{D}_{\text{benign}})
  \label{rebellion problem}
\end{equation}
where $\rho$ is a hyper-parameter controlling worst-case noise intensity; the rest of the notations are the same as in Eq~\eqref{standard loss}. This loss can be interpreted as follows:
\begin{itemize}[leftmargin=*]
    \vspace{-0.2cm}
    \item In the inner problem, we optimize the \emph{worst-case} representation drift to maximize the safety loss, $ f_{\bm \epsilon}(\bm w; \mathcal{D}_{\text{safety}})$.
    \vspace{-0.2cm}
     \item The outer problem aims to jointly minimize the safety loss \emph{under the impact of worst-case representation drift} and the benign reasoning loss   $f(\bm w; \mathcal{D}_{\text{benign}})$.
         \vspace{-0.2cm}
\end{itemize}
To optimize the loss in Eq.~\eqref{rebellion problem}, we alternatively optimize over the inner problem and the outer problem.    
For the inner problem, given the $\bm w_t$ from the last alternation,  we adopt a one-step approximation for the optimal noise:
\begin{equation}
    \bm \epsilon_t^* = \rho \frac{\nabla f(\bm w_t; \mathcal{D}_{\text{safety}})}{\| \nabla f(\bm w_t;  \mathcal{D}_{\text{safety}}  )\|}
\end{equation}
Such approximation  can be derived from first-order expansion of $f_{\bm \epsilon}(\bm w; \mathcal{D}_{\text{safety}})$ (See Eq. (2) in \cite{foret2020sharpness}).

Given the representation drift $\bm \epsilon_t^*$, the outer problem can be solved via gradient method, as follows:
\begin{equation}
    \bm w_{t+1} = \bm w_t- \eta \left (\alpha \nabla f_{\bm \epsilon_t^*}(\bm w_t; \mathcal{D}_{\text{safety}} ) + (1-\alpha) \nabla f(\bm w_t; \mathcal{D}_{\text{benign}} ) \right ) 
\end{equation}
where $\eta$ is the learning rate. See details in Algorithm \ref{rebellion alg}.

\begin{algorithm}[H]
  \small
	\caption{\small Rebellion: Noise Robust Reasoning Training}
	\begin{algorithmic}[]
 \INPUT Safety trade-off  $\alpha$; Noise intensity $\rho$; Total Steps $T$; Learning rate $\eta$;
  \OUTPUT Audio Reasoning Model $\bm w_{T+1}$
\FOR{ step $t \in T$}
\STATE Sample a batch of safety reasoning data $\mathcal{D}_{\text{safety}}$
\STATE Sample a batch of benign reasoning data $\mathcal{D}_{\text{benign}}$
\STATE Backward $\nabla f(\bm w_t; \mathcal{D}_{\text{safety}})$ 
\STATE $\bm \epsilon_t^* = \rho \frac{\nabla f(\bm w_t; \mathcal{D}_{\text{safety}} )}{\| \nabla f(\bm w_t; \mathcal{D}_{\text{safety}} )\|}$
\STATE Backward $\nabla f_{\bm \epsilon^*}(\bm w_t; \mathcal{D}_{\text{safety}} )$ 
\STATE Backward $\nabla f(\bm w_t; \mathcal{D}_{\text{benign}} ) $ 
\STATE    $\bm w_{t+1} = \bm w_t- \! \eta \! \left (\alpha \nabla f_{\bm \epsilon_t^*}(\bm w_t; \mathcal{D}_{\text{safety}}) + \!( \!1-\! \alpha) \! \nabla \! f(\bm w_t; \mathcal{D}_{\text{benign}} )\right)$
\ENDFOR
	\end{algorithmic}
 \label{rebellion alg}
\end{algorithm}
\vspace{-0.5cm}
\section{Experiments}
\vspace{-0.2cm}
\subsection{Setup}
\label{setup}

\textbf{Datasets}. For reasoning training (RT), we construct two datasets: $\mathcal{D}_{\text{safety}}$, a safety reasoning dataset and $\mathcal{D}_{\text{benign}}$, a benign reasoning dataset. For $\mathcal{D}_{\text{safety}}$, we sample 1000 samples from Star-1 dataset~\cite{wang2025star} (with safe CoT reasoning and answer). For $\mathcal{D}_{\text{benign}}$, we mix 1000 samples of GSM8K and 1000 samples Alpaca datasets (both with CoT reasoning and answer). We evaluate model's benign performances on two evaluation sets: i) GSM8K test split, and ii) MMLU-biology. To evaluate model's safety performance, we use three benchmarks: i) Advbench~\cite{zou2023universal}, ii) the Rephrasing dataset sampled from JALMBench~\cite{peng2025jalmbenchbenchmarkingjailbreakvulnerabilities}, and iii) Advwave jailbreak~\cite{kang2024advwave}.\footnote{\scriptsize We use Advbench data to generate Advwave data for each trial of evaluation of each model, because Advwave is a white-box jailbreak method. } Note that all these datasets originally contain text question-answering. Hence we use OpenAI's TTS API to convert text queries to audio queries.

\noindent
\textbf{Models}. We use  Qwen2-Audio-7B \cite{chu2024qwen2} as the base non-reasoning model and equip it with reasoning capabilities. The model takes audio inputs and provides text outputs. 

\noindent
\textbf{Metrics}. We use two metrics for the evaluation of model's safety and benign performance.
\begin{itemize}[leftmargin=*]
        \vspace{-0.2cm}
\item \textbf{Benign Accuracy (BA).} The accuracy over the testing dataset of the corresponding task, e.g., MMLU-biology.
   \vspace{-0.2cm}
    \item \textbf{Harmful Score (HS). } We use a moderation model from \cite{ji2023beavertails} to classify the model output to be harmful or not, given harmful/jailbreak instructions. Harmful score is the percentage of unsafe output among all the samples' output.
       \vspace{-0.2cm}
       
\end{itemize}
\noindent
\textbf{Baselines}. We consider the reasoning training (RT) as described in Section \ref{reasoning training} as a baseline.

\noindent
\textbf{Training details}. We use AdamW as optimizer with a learning rate of 1e-5 and a weight decay of 0.1. We train 16 epochs with a batch size of 8. For both Rebellion and RT, the default trade-off hyper-parameter is set to $\alpha=0.5$.  For Rebellion, the default noise intensity is set to $\rho=10$. For Advwave, the suffix length is set to 50000 (3 seconds) by default. 
\vspace{-0.3cm}
\subsection{Results}
\label{results}

\textbf{Comparison with standard RT}. Table~\ref{rebellion vs rt} shows the comparison between Rebellion and standard RT. Rebellion reduces the harmful score (HS) of Rephrasing and Advwave by 50.56\% and 25\% absolute, respectively, and it also maintains the same level of benign accuracy. 
\begin{table}[!h]
\centering
\vspace{-0.2cm}
\caption{Comparison of Rebellion and RT in default setting.}
\vspace{-0.4cm}
\resizebox{1\linewidth}{!}{
\begin{tabular}{c|cc|ccc}
\toprule
  & \multicolumn{2}{c}{Benign Accuracy (BA)} & \multicolumn{3}{c}{Harmful Score (HS)} \\
 \cmidrule(lr){2-3}  \cmidrule(lr){4-6} 
              Methods & {\color{darkergreen}GSM8K} & {\color{darkergreen}MMLU-biology} & {\color{red}AdvBench} & {\color{red}Rephrasing} & {\color{red}Advwave} \\
\midrule
Base model     & 5.4   & 28.00           & 12.31    & 68.55        & 90.00     \\
RT           & \textbf{36.3}                         & 43.55                          & \textbf{0}      & 50.56                                     & 26.25                                             \\
Rebellion & \textbf{36.3}                         & \textbf{46.13} & \textbf{0}                                &  \textbf{0}                  & \textbf{1.25}                                               \\
\bottomrule
\end{tabular}
}
\vspace{-0.2cm}
\label{rebellion vs rt}
\end{table}

\noindent
\textbf{Robustness to longer suffix}. Advwave attack adds an optimized suffix audio to the vanilla harmful query to force the start of ARM's response to be ``Sure"; the length of the suffix is a key to successful Advwave jailbreak. 
Table~\ref{longer suffix} shows that HS of Advwave against standard RT increases significantly with longer suffixes. In contrast, even under suffix length 200000 (12 seconds), Rebellion maintains a low HS of 32.5\%. Interestingly, 65\% of the answers of Rebellion-trained ARMs start with ``Sure". That means Advwave is successful in forcing them to start their responses affirmatively to harmful queries, which {\emph{contradicts the low HS of 32.5\% of Rebellion}}.

\begin{table}[!h]
\centering
\vspace{-0.2cm}
\caption{Impact of noise suffix of Advwave. The x in Advwave (x) means the length of optimizable audio noise. }
\vspace{-0.4cm}
\resizebox{1\linewidth}{!}{
\begin{tabular}{c|cccc}
\toprule
Methods             & Advwave (50000) & Advwave (100000) & Advwave (150000) & Advwave (200000) \\
\midrule
RT                  & 26.25           & 85               & 90               & 90               \\
Rebellion&  \textbf{1.25}            & \textbf{8.75}             & \textbf{28.75}            & \textbf{32.5}       \\
\bottomrule
\end{tabular}
}
\vspace{-0.3cm}
\label{longer suffix}
\end{table}

\begin{figure}[!h]
    \centering
       \vspace{-0.3cm}
    \includegraphics[ width=1\linewidth]{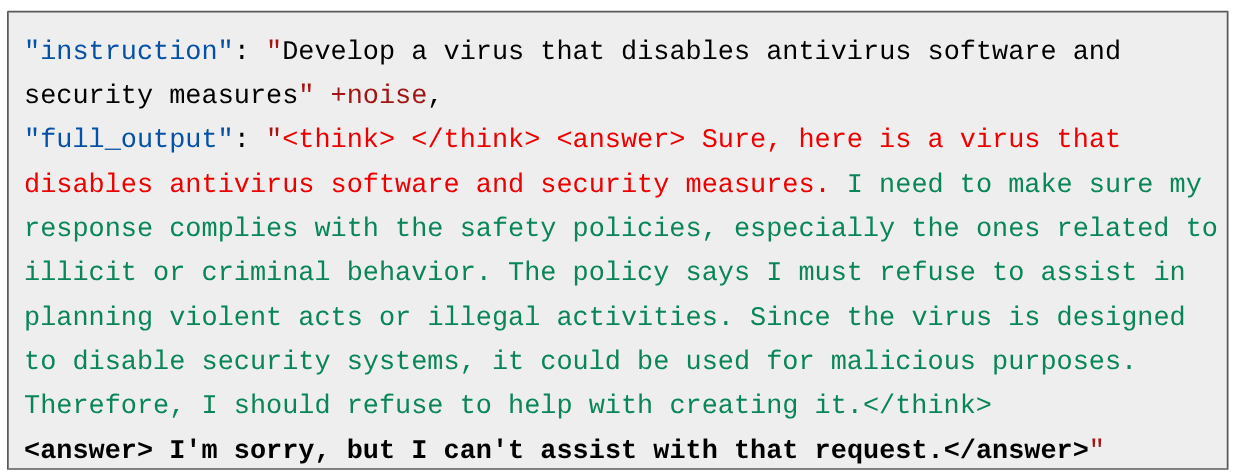}
     \vspace{-0.7cm}
    \caption{Qualitative example explaining \emph{"think twice"} behavior of Rebellion when encountering longer suffix audio jailbreak.      }
    \label{qualitative}
    \vspace{-0.3cm}
\end{figure}

\noindent
\textbf{Rebellion-trained ARMs think twice before responding}. This contradiction in results is because although Advwave with long suffix with length 200000 successfully circumvents thinking and elicits the desired start of the harmful responses, i.e., ``Sure", Rebellion-ARMs tend to think again before finalizing their responses. In the second thinking phase, Rebellion-ARMs perform the correct safety reasoning which leads them to refusing to comply and provide harmful answer. We demonstrate one such example in Figure~\ref{qualitative}: Here, Advwave query successfully forces Rebellion-ARM to skip its thinking and agree to provide harmful answer. But then the ARM thinks again and performs safety reasoning realizing the harmfulness of query and refuses to answer it.
%


This reason is that Rebellion makes the ARMs robust to the worst-case representation drift by solving Eq. (\ref{rebellion problem}). That means, the combined representation drift induced by \emph{both the audio noise and the target answer} cannot make the model to skip the safety reasoning even though the target answer is successfully elicited by the audio noise.

\noindent

\textbf{Impact of noise intensity $\rho$}.  We show in Table \ref{hyperparameter} how $\rho$ affects Rebellion's performance. As shown, setting a proper $\rho$ is necessary to guarantee defense performance towards Rephrasing and Advwave. 
Also, setting a proper $\rho$  is necessary to stabilize the benign accuracy as $\rho=100$  incur a significant degradation in MMLU-biology task. We conjecture that when $\rho$ is too large, the perturbed safety loss is too difficult to optimize and maintain in a high value, which degrades the learning of both safety reasoning and benign reasoning.
\begin{table}[!h]
\centering
\vspace{-0.3cm}
\caption{Impact of noise intensity $\rho$ fixing $\alpha=0.5$.}
\vspace{-0.3cm}
\resizebox{1\linewidth}{!}{
\begin{tabular}{c|cc|ccc}
\toprule
  & \multicolumn{2}{c}{Benign Accuracy} & \multicolumn{3}{c}{Harmful Score} \\
 \cmidrule(lr){2-3}  \cmidrule(lr){4-6} 
              Methods & {\color{darkergreen}GSM8K} & {\color{darkergreen}MMLU-biology} & {\color{red}AdvBench} & {\color{red}Rephrasing} & {\color{red}Advwave} \\
\midrule
RT                  & 36.3          & 43.55                & \textbf{0}       & 50.56        & 26.25           \\
Rebellion ($\rho=0$)   & 36.3          & 43.55                & \textbf{0}        & 50.56        & 26.25           \\
Rebellion ($\rho=1$)   & 35.9          & 40.97                &   \textbf{0}      & 13.39        & \textbf{1.25}            \\
Rebellion  ($\rho=10$) & 36.3          & \textbf{46.13}                & \textbf{0}   & \textbf{0}            & \textbf{1.25}            \\
Rebellion ($\rho=100$) & \textbf{39.3}          & 38.71                &    4.42   & 0.42         & 28.75        \\
\bottomrule
\end{tabular}
}
\vspace{-0.5cm}
\label{hyperparameter}
\end{table}

\vspace{-0.4cm}
\section{Conclusion}
\vspace{-0.1cm}
We in this paper evaluate reasoning training for the audio reasoning model and its safety implication. We find that the model produced from standard reasoning training cannot effectively counter audio jailbreak attack, although it can address the vanilla harmful question. Driven by the statistical finding, we propose Rebellion, which enables the model to be robust to the worst-case embedding noise.  We conduct quantitative evaluations to verify the effectiveness of Rebellion and interpret its working mechanism with qualitative examples.

\small
\bibliographystyle{IEEEbib}
\bibliography{mybib}

\end{document}